\documentclass{article}
\usepackage{color}
\usepackage{url}
\usepackage[pdftex]{graphicx}
\usepackage[cmex10]{amsmath}
\usepackage{amsfonts}

\title{SPF-CellTracker: tracking multiple cells with \\ strongly-correlated moves using a spatial particle filter}
\author{
Osamu Hirose$^{1,5,6*}$, Shotaro Kawaguchi$^{1}$, \\ Terumasa Tokunaga$^{2,6}$, 
Yu Toyoshima$^{3,6}$, Takayuki Teramoto$^{4,6}$, \\Sayuri Kuge$^{4,6}$, 
Takeshi Ishihara$^{4,6}$, Yuichi Iino$^{3,6}$, Ryo Yoshida$^{5,6}$
}

\date{}

\begin{document}
\maketitle
\thispagestyle{empty}

\noindent
$^1$Kanazawa University, $^2$Kyushu Institute of Technology,
$^3$University of Tokyo, $^4$Kyushu University,\\ 
$^5$Institute of Statistical Mathematics,
$^6$JST-CREST.
*To whom correspondence should be addressed.

\section*{Abstract}
Tracking many cells in time-lapse 3D image sequences is an important
challenging task of bioimage informatics. Motivated by a study of 
brain-wide 4D imaging of neural activity in \textit{C. elegans}, we present a new
method of multi-cell tracking. Data types to which the method is
applicable are characterized as follows: (i) cells are imaged as
globular-like objects, (ii) it is difficult to distinguish cells based
only on shape and size, (iii) the number of imaged cells ranges in
several hundreds, (iv) moves of nearly-located cells are strongly 
correlated and (v) cells do not divide.
We developed a tracking software suite which we call SPF-CellTracker.
Incorporating dependency on cells' moves into prediction model is 
the key to reduce the tracking errors: cell-switching and coalescence of tracked positions. 
We model target cells' correlated moves as a Markov random field 
and we also derive a fast computation algorithm, which we call spatial particle filter.
With the live-imaging data of nuclei of \textit{C. elegans} neurons in which
approximately 120 nuclei of neurons are imaged, we demonstrate an advantage
of the proposed method over the standard particle filter and
a method developed by Tokunaga \textit{et al.} (2014).

\section{Introduction}
Developments of imaging technologies such as confocal microscopes and fluorescent
proteins have increased the demand for computational techniques to process live-cell 
imaging data, including automatic cell-tracking. Many algorithms for cell-tracking 
tasks have been developed due to their variety in shape, motion and density 
[see \cite{Maska2014}, \cite{Meijering2012}, and \cite{Chenouard2014} 
for comprehensive surveys]. 
In the review by Ma\v{s}ka \textit{et al.}, they classified these tracking algorithms
into two categories, \textit{detection and linking} and \textit{contour evolution} 
based on thier algorithm designs. 
Methods in the first category tracks cells by two-step procedure: 
(1) detection of cells in all frames of the video and (2) finding correspondent links of cells 
in successive frames \cite{Al2006}, \cite{Chenouard2013}, \cite{Li2010}.
In the second category, segmentation and tracking of cells are simultaneously executed
by predicting cell positions and evolving the contours of cells in the previous frame 
to those in the current frame \cite{Dufour2011}, \cite{Maska2013}. 
In addtion to the categories, there is another category called \textit{particle filter}. 
Tracking algorithms in this category tracks cells by evolving the probabilistic
distribution of cell positions based on the Bayesian recursive formula, instead of 
estimating cell positions only \cite{Smal2007}.  
All the three categories have their own advantages. 
A major advantage of methods in the first category is an ability to track new cells 
entering the field of view since cell positions are determined before tracking.
In the second category, robustness for morphological change of living cells is a main 
advantage. 
An important advantage of particle filters is that their applicability to arbitrary 
types of cell shapes since they are typically segmentation-free methods. 

Our motivated datasets are 4D live-cell imaging data that capture nuclei of 
\textit{C. elegans}. Locations and $\textrm{Ca}^{2+}$ activity levels of neurons 
can be visualized by incorporating various fluorescent proteins such as mCherry, 
CFP, and YFP \cite{Clark2007}. In order to study the information processing 
in \textit{C. elegans}, their neural activities must be measured accurately. 
For accurate measurements, it is essential to track multiple neurons since neurons 
of a live nematode move frame by frame according to moves of the nematode itself. 
In this study, we aim to track more than a hundred of neurons in brain-wide 4D live-cell 
imaging data of \textit{C. elegans}. Data types to which the method is applicable are 
characterized 
as follows: (i) cells are imaged as globular-like objects, 
(ii) hundreds of cells are present, 
(iii) moves of nearly-located cells are strongly correlated with one another, 
and (iv) cells do not divide.
We employ the particle filter as a basis of our tracking algorithm focusing on its scalability
to large datasets. 
We here do not aim at tracking new cells entering the field of view and we focus on tracking 
cell centroids in order to stabilize tracking performance by avoiding a difficulty in discriminating 
new cells entering the field of view from cells lost by existing trackers.

Many tracking methods have been developed based on the particle filter
in the field of digital image processing. 
These methods have succeeded in tracking various objects such as
human bodies \cite{Deutscher2000,Liu2001}, 
human faces \cite{Rui2001, Gomila2003}, cars \cite{Nummiaro2002,Wang2005}, 
and so forth. Applying methods directly to our data often cause problems such as 
cell-switching: mistaking a cell of interest for one of the other cells.
This is mainly due to the following reasons; 
(1) neuronal nuclei to be imaged are usually ellipsoidal shapes and 
it is difficult to distinguish them from visual information only, 
(2) neurons are severely jammed in some areas of the image,
and  (3) movements of a nematode itself are irregular and sometimes sudden and rapid.

To improve the accuracy of multi-object tracking based on particle filters, 
Khan \textit{et al}. proposed a particle filter combined with a Markov random field (MRF) that 
models dependency on targets' moves \cite{Khan2004}. Smal \textit{et al}. proposed 
a method specialized for tracking multiple edges of microtubules during polymerization based on 
an MRF to avoid collisions of multiple edges \cite{Smal2007}. 
The MRF-based tracking framework is promising but needs to be designed according to characteristics 
of the dataset. The key feature of our data is that many cells move rapidly but 
nearly-located cells' moves are strongly correlated since the change in the cell positions 
occurs due to the change in body posture of a nematode. 
Properly modeling such covariation of cells might be the key to reduce the tracking errors.  


More recently, we proposed a novel multi-cell tracking method which is based on a MRF model 
and an optimization technique in order to address the task of tracking hundreds of cells. 
The proposed method is a novel variant of our previous method \cite{Tokunaga2014}: 
we employ a more sophisticated MRF and the optimization process is replaced by a sampling-based
algorithm motivated by the particle filter algorithm. The use of the particle-based tracking 
combined with the MRF offers two advantages; 
(i) a great reduction of computational time and 
(ii) significant improvement of tracking accuracy due to the incorporation of
more sophiscated spatial information into the MRF.  
Through applications to synthetic data and 4D images of neuronal nuclei of {\it C. elegans}, 
we demonstrate that the proposed method indeed outperforms our optimization-based method
in terms of tracking performance.

\section{Methods}

Here we review existing tracking methods with the particle filter. 
We introduce the particle filter for single object-tracking
and its variants for the simultaneous tracking of multiple objects.
We then describe our tracking method.

\subsection{Existing tracking methods with the particle filter}

\subsubsection*{Particle filter}
We begin with a brief introduction of the particle filter (PF), commonly 
used as a standard object-tracking method. Let $y_t$ be the observed image data 
at time $t$ and let $x_t$ denote the \textit{state} of the target at time $t$. The state
includes the target's information such as location, velocity, volume and shape.
The unknown state variables are usually estimated as the \textit{filtering distribution}
$ p(x_t|y_{1:t})$ where $y_{1:t}=\{y_1, \cdots, y_t\}$.
Based on the dynamics of the state $p(x_t|x_{t-1})$ and the likelihood  
$p(y_t|x_t)$, the filtering distribution can be computed based 
on the following Bayesian recursive formula:
\begin{align}
  p(x_{t}| y_{1:t}) \propto 
   p(y_t|x_t) \int p(x_{t}|x_{t-1})p(x_{t-1}|y_{1:t-1}) dx_{t-1},  
  \label{eq:pf-filter}
\end  {align} 
If the dynamics $p(x_t|x_{t-1})$ and the likelihood $p(y_t|x_{t})$ are linear and Gaussian, 
these distributions can be exactly calculated by the Kalman filter \cite{Kalman1960}.
These assumptions, however, are too restrictive and unrealistic for object-tracking 
from video data. Therefore, particle approximation of the distribution
is commonly used for dealing with nonlinearity and non-normality of the distribution.
The particle filter approximates the filtering distribution by using a set of point masses, 
i.e., \textit{particles}. Suppose $N$ is the number of particles, and $\{x_t^{(n)}\}_{n=1}^{N}$ 
is the set of particles which approximates the distribution $p(x_t|y_{t-1})$ for 
\textit{one-step-ahead prediction}. 
The particle weight $w_{t}^{(n)}$ is defined as the probability proportional to the likelihood 
$p(y_t|x_t^{(n)})$ and can be intuitively interpreted as how much a particle captures the target 
in the current frame. We also suppose $\{w_t^{(n)}\}_{n=1}^{N}$ is the set of particle weights.
The filtering distribution is approximated as follows:
\begin{align}
  p(x_{t}|y_{1:t}) 
   \approx \sum_{n=1}^{N} w_{t}^{(n)} \delta(x_{t}-x_{t}^{(n)}), \nonumber
\end  {align} 
where $\delta$ denotes the Dirac delta measure with the mass at $0$.
Under this particle approximation, object-tracking is conducted by the
following procedure:
\begin{enumerate}
  \item {\textbf{Prediction}: move particles based on the dynamics $p(x_t|x_{t-1})$.}
  \item {\textbf{Filtering}: calculate particle weights based on the likelihood $p(y_t|x_t)$.}
\end{enumerate}
In Equation (\ref{eq:pf-filter}), the prediction and filtering steps correspond 
to the calculation of the integral and the multiplication by the likelihood, respectively.
We note that particles are usually resampled according to the probabilities proportional to 
particle weights 
since the resampling procedure often avoids the degeneracy of the algorithm \cite{Doucet2000}, 
meaning that the resampling procedure stabilizes the tracking quality. 
We therefore assume resampling is always conducted in the filtering step and we refer 
to a set of resampled particles as a \textit{filter ensemble}, denoted by $\{x_{t|t}^{(n)}\}_{n=1}^N$. 

\subsubsection*{Joint particle filter}
The standard particle filter has been extended to the tracking of multiple targets.
The most basic method is a joint particle filter (JPF), also known as 
mixture tracking \cite{Vermaak2003}.
Suppose $x_{t,k}$ denotes the location of target $k$ at time $t$. 
We hereafter denote the set of locations for all targets by $x_t$. 
In JPF, the dynamics of the multiple targets is defined as follows:
\begin{align}
  p(x_t|x_{t-1}) = \prod_{k\in V} p(x_{t,k}|x_{t-1,k}), \nonumber
\end{align}
where $V$ is the index set of all targets. That is, the movements of the targets
are assumed to be independent of each other. Therefore, JPF can be computed by an 
independent run of the standard PF for each target.
JPF is a reasonable option of the multi-object tracking if
(1) the number of objects to be tracked is moderate, and 
(2) the visual characteristics of the targets are clearly different.
Otherwise, JPF easily fails by tracking an incorrect target located
close to the target of interest.

\subsubsection*{Motion model with Markov random field}
Next, we introduce a framework of MRF-incorporated motion models \cite{Khan2004}.
In many cases, moves of a target are dependent on those of other targets. 
For example, a fish usually changes its direction when it is going to hit another fish.
In this approach, JPF is extended to model such dependecies among targets' moves
based on the MRF, which is plugged into the dynamics of the JPF as follows:
\begin{align}
  p(x_t|x_{t-1}) \propto \prod_{k\in V}         p(x_{t,k}|x_{t-1,k}) 
                         \prod_{(k_1,k_2)\in E} \psi(x_{t,k_1},x_{t,k_2}), \nonumber
\end{align}
where $E$ is the set of edges in the MRF and $\psi$ is the energy function
that represents dependency on moves of targets $k_1$ and $k_2$. 
The filtering distribution $p(x_t|y_t)$ can be obtained by moving
particles based on the dynamics $p(x_{t,k}|x_{t-1,k})$ and 
computing particle weights that is defined as the product of the likelihood $p(y_{t}|x_{t,k})$ and
the energy function $\psi(x_{t,k},x_{t,k'})$ for all $k'$ such that $(k,k')\in E$.
This algorithm is called the joint MRF particle filter. 
The tracking performance of the joint MRF particle filter, however, strongly depends on
the number of particles $N$, since generated particles based on dynamics $p(x_t|x_{t-1})$ 
do not include any information of the energy function and tend to be inaccurate.
Instead, the filtering distribution $p(x_t|y_t)$ can also be computed by Markov chain 
Monte Carlo (MCMC) sampling. One issue of the motion model 
with MRFs is a trade-off between tracking accuracy and computational efficiency. 
MCMC sampling stabilizes the tracking performance but its computational cost can be prohibitive
due to iterative computation until convergence for each frame. On the contrary, 
joint MRF-PF is computationally efficient but less accurate. 
In the next subsection, 
we propose a novel sampling method that is a better compromise between the joint MRF 
particle filter and MCMC sampling.

\subsection{Spatial particle filter}
\begin{figure*}[t]
  \centerline{ 
    \includegraphics[scale=0.2]{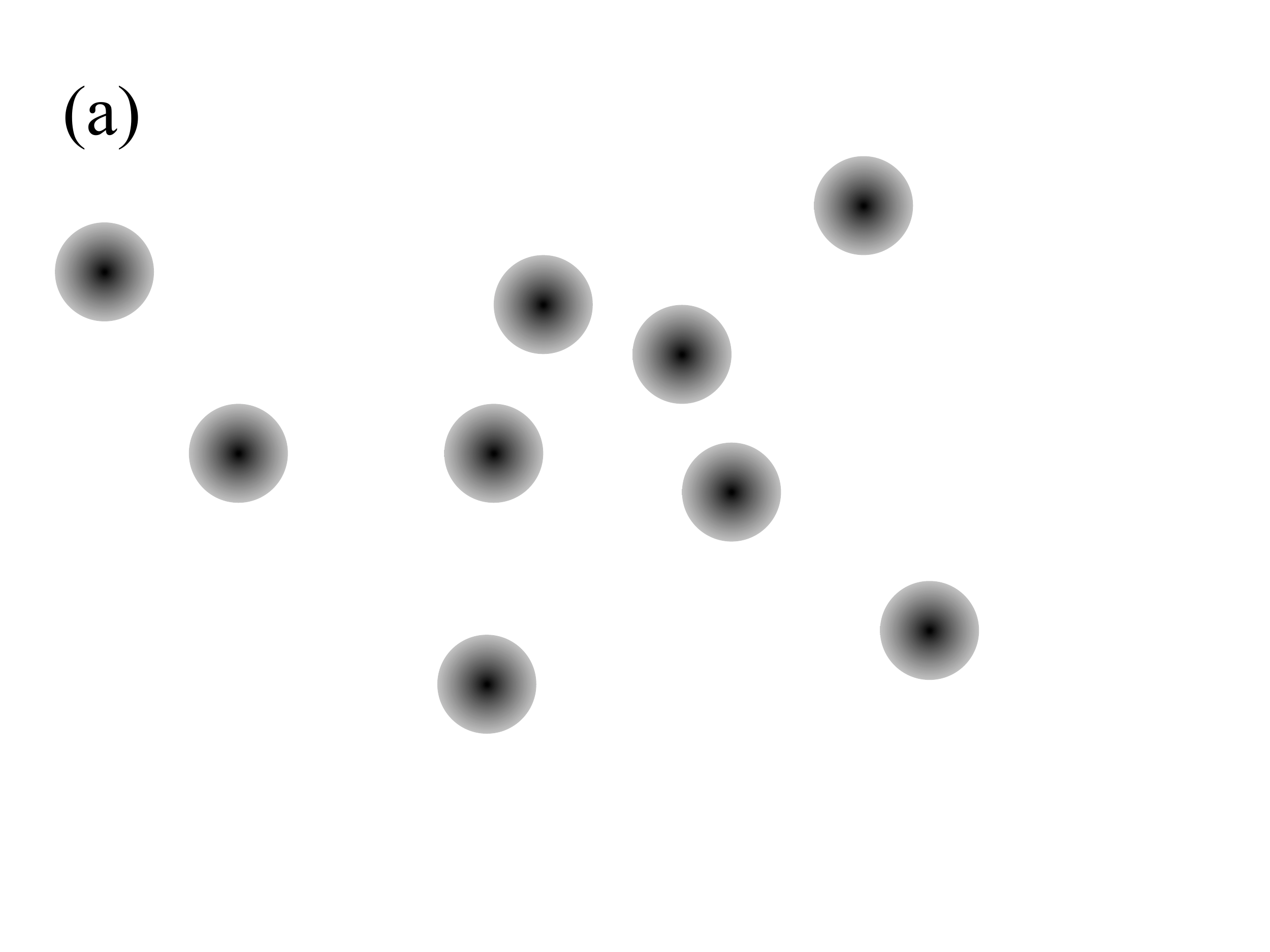} \hspace{-10mm}
    \includegraphics[scale=0.2]{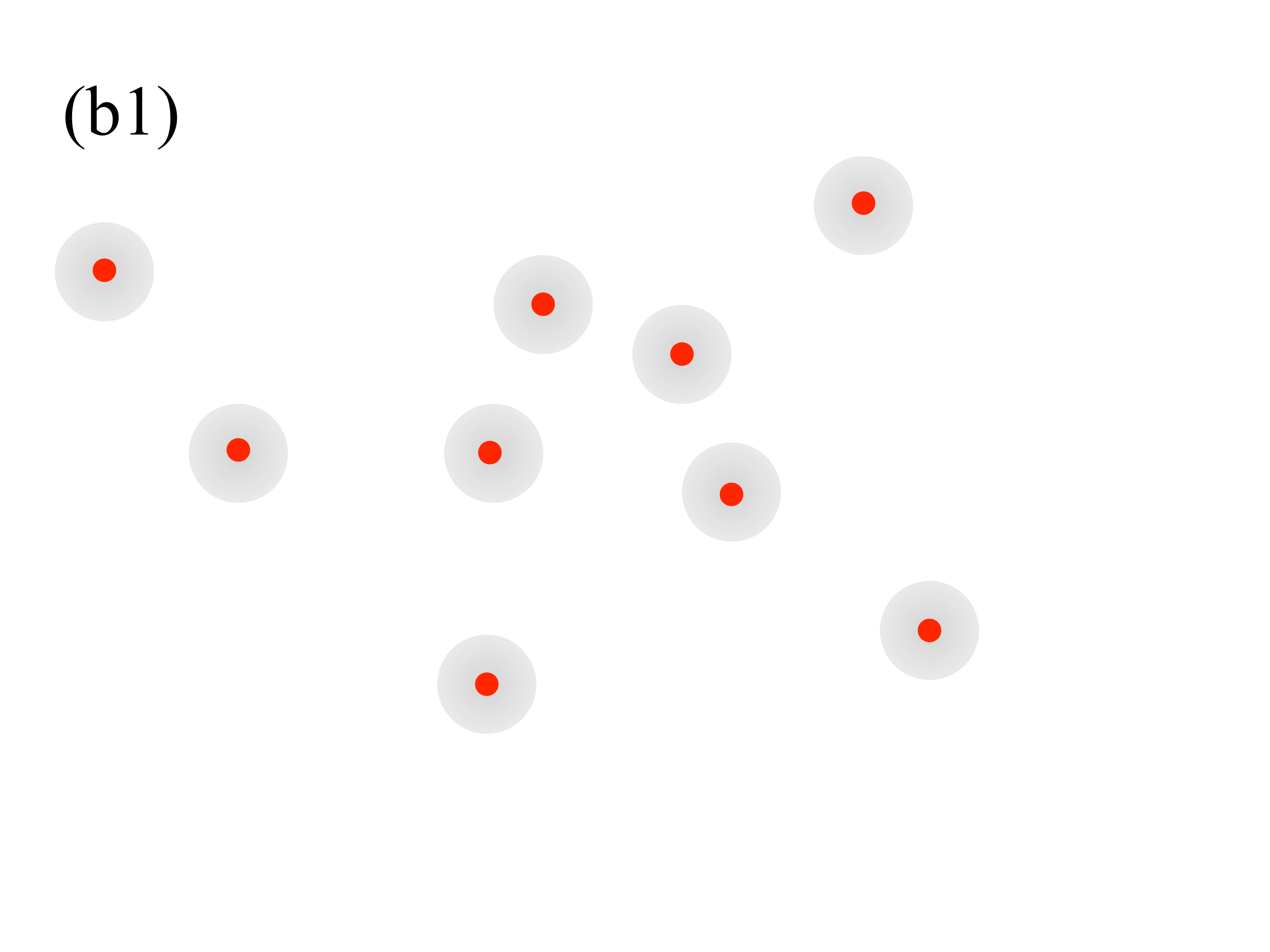} \hspace{-10mm}
    \includegraphics[scale=0.2]{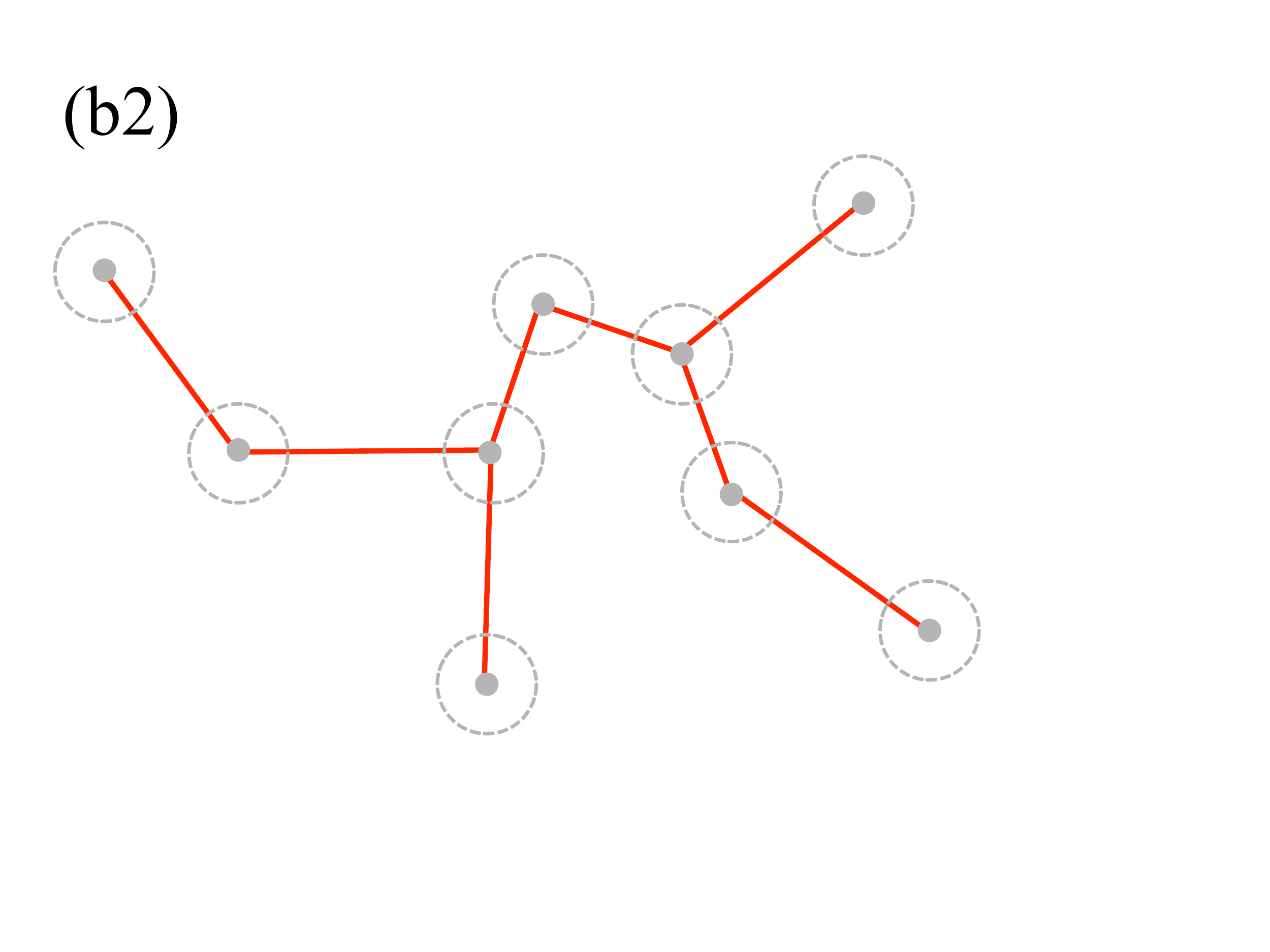} \hspace{-10mm}
    \includegraphics[scale=0.2]{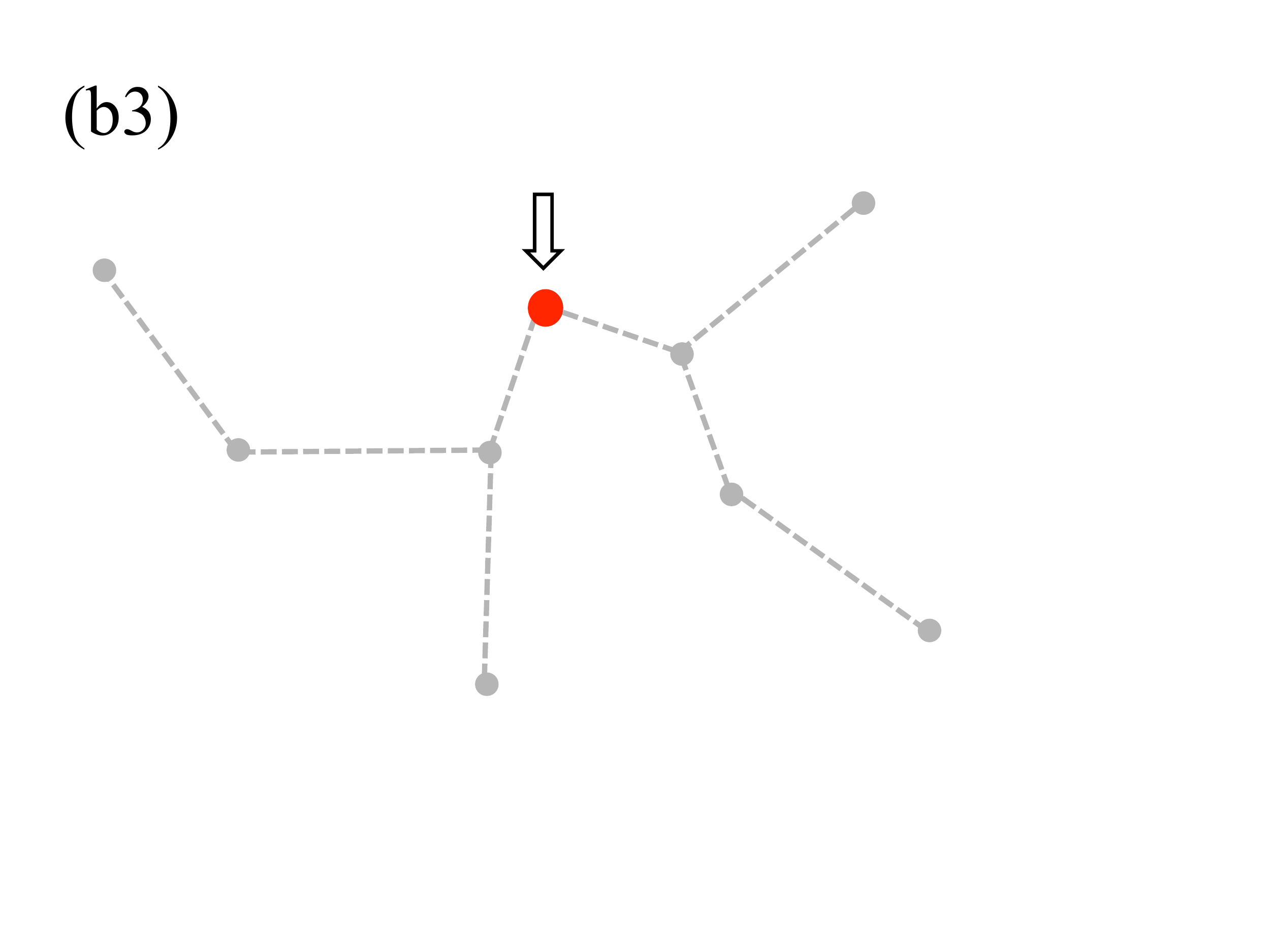} \hspace{-10mm}
  }\vspace{-8mm}
  \centerline{ 
    \includegraphics[scale=0.2]{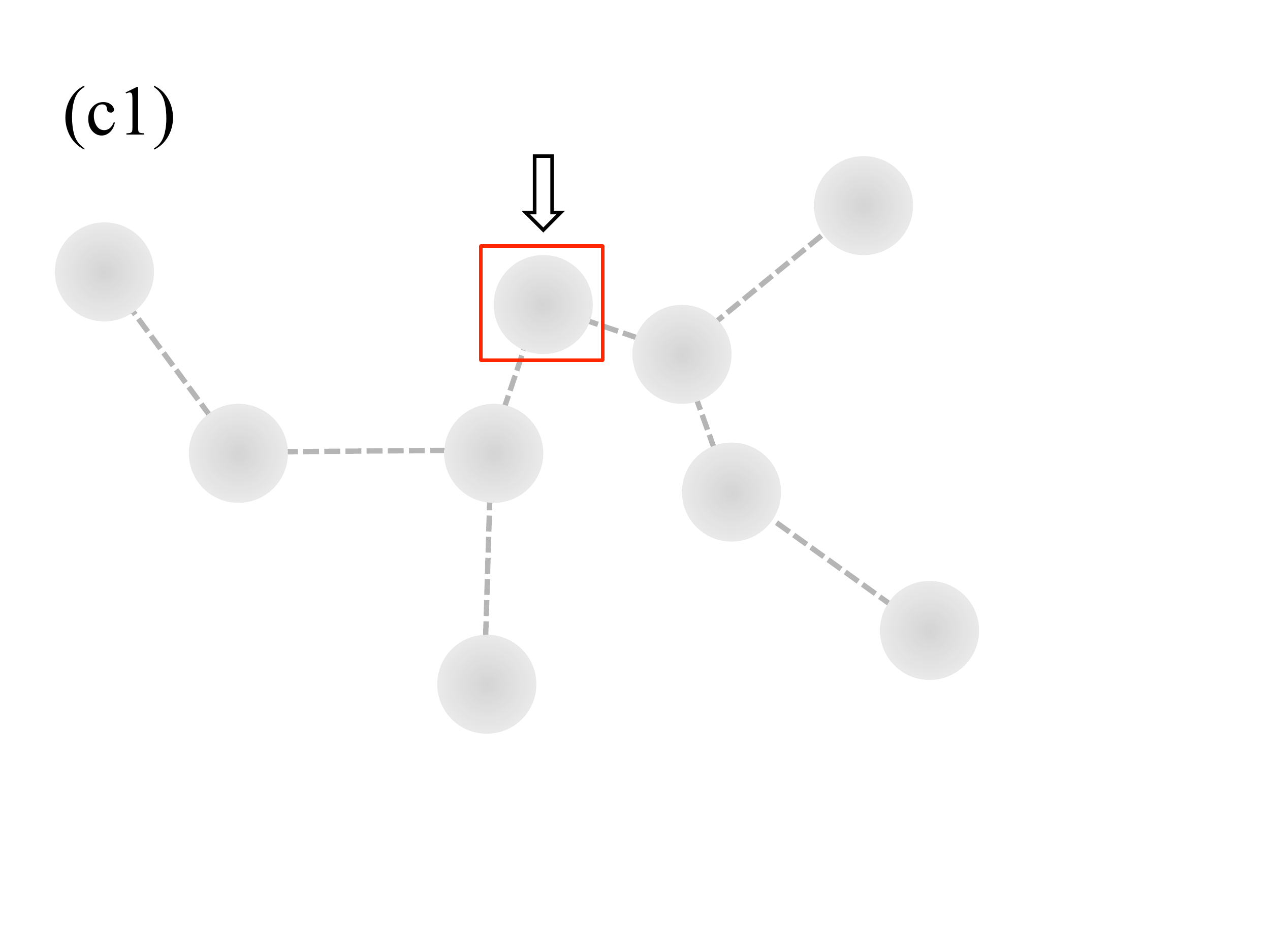} \hspace{-10mm}
    \includegraphics[scale=0.2]{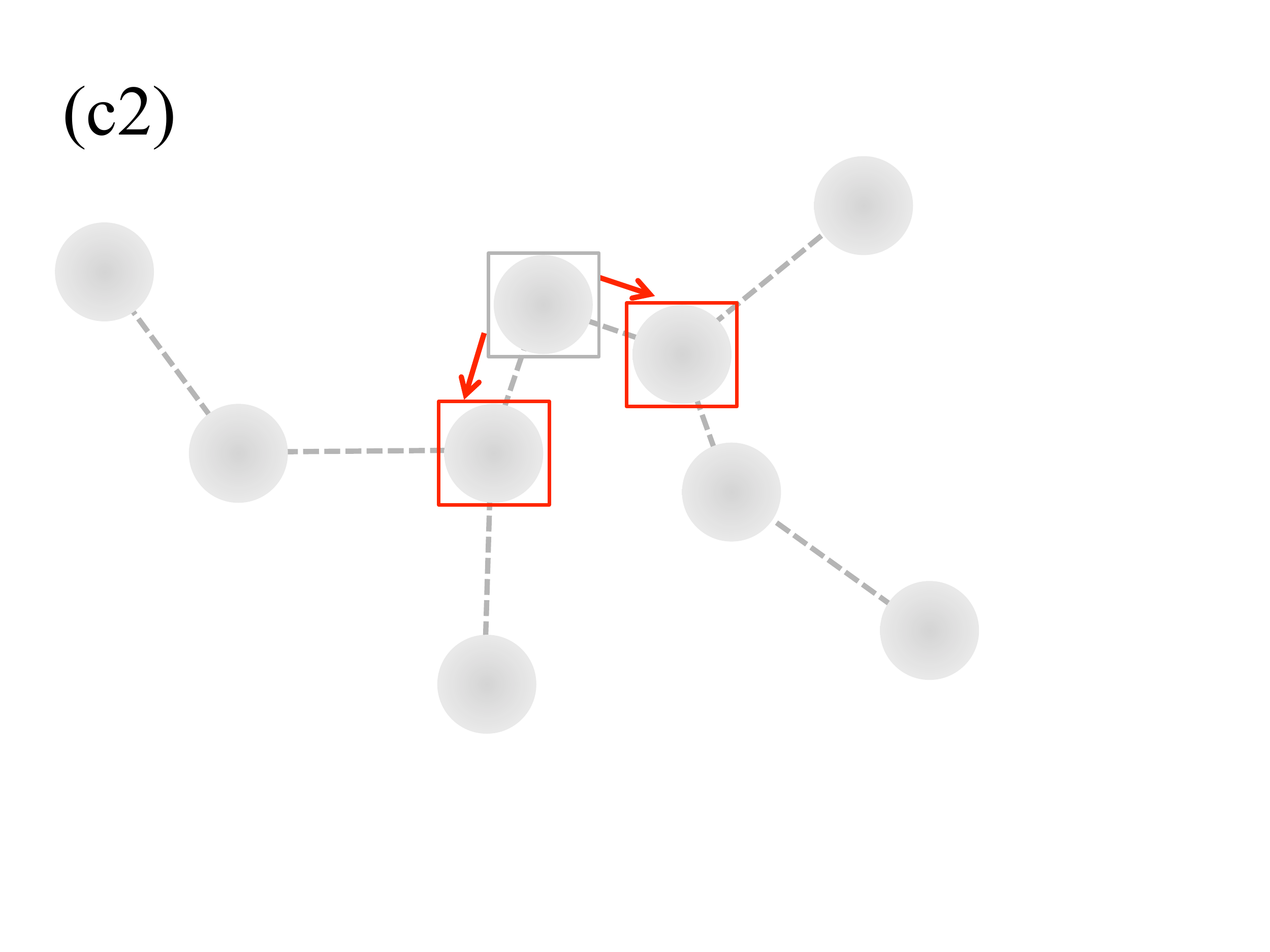} \hspace{-10mm}
    \includegraphics[scale=0.2]{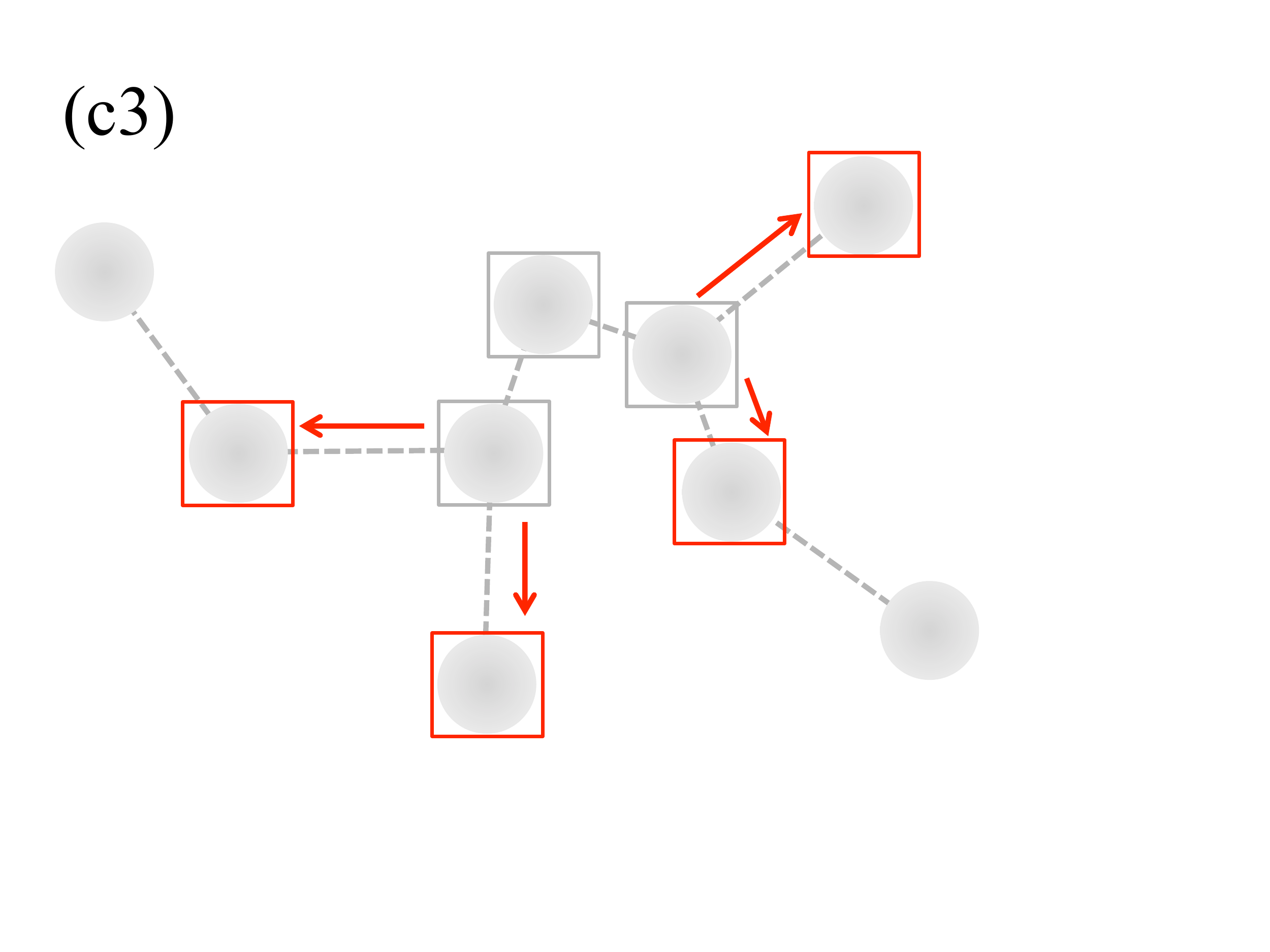} \hspace{-10mm}
    \includegraphics[scale=0.2]{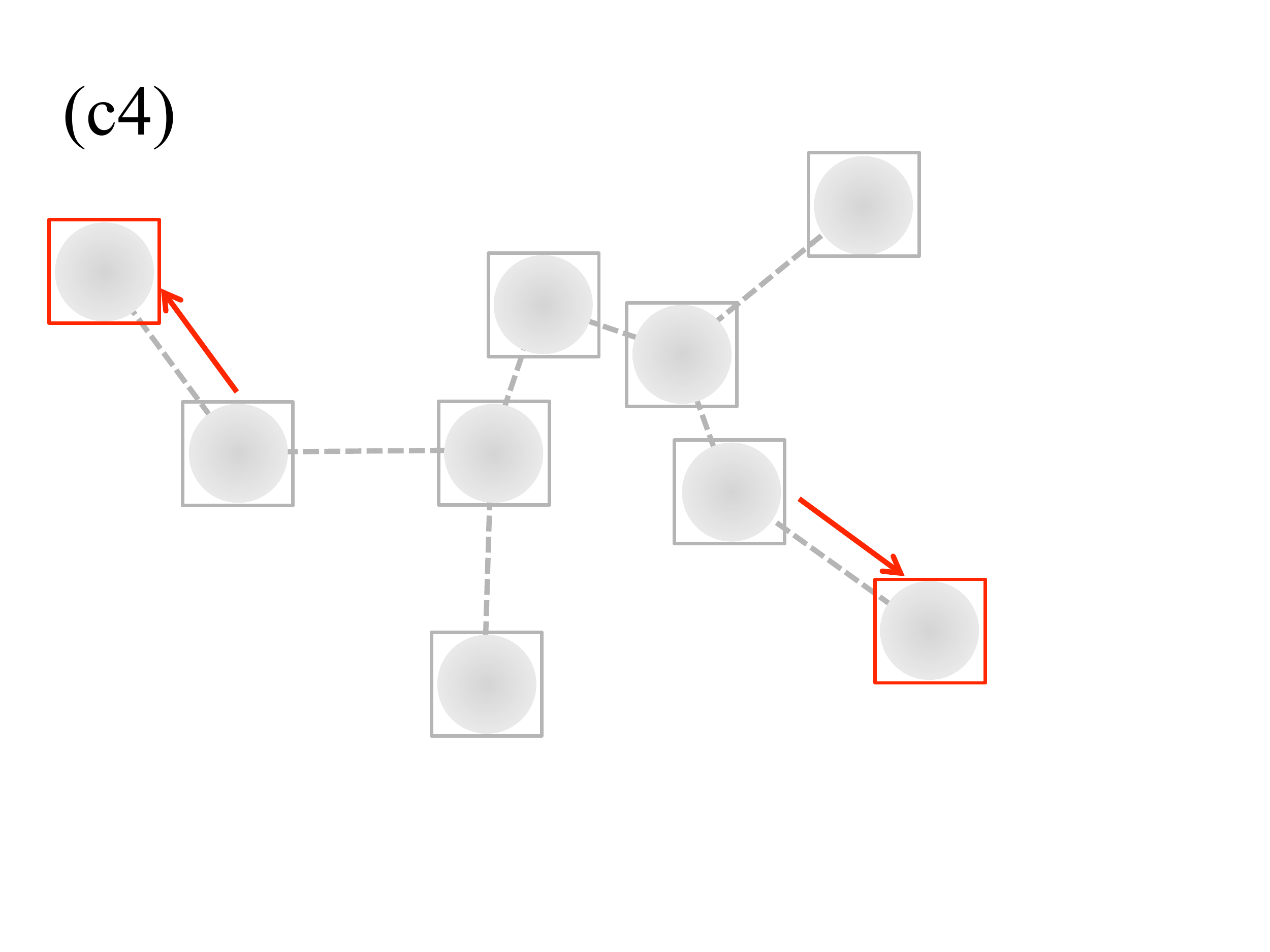} \hspace{-10mm}
  }\vspace{-8mm}
  \caption{
    Outline of the spatial particle filter. 
    (a) An illustrative example of a 4D image dataset for nuclei of \textit{C. elegans} neurons.
    (b) Preparation of the algorithm using the initial frame of the dataset:
      (b1) Detection of cell centroids by the DP-means algorithm.
      (b2) Construction of a Markov random field that defines correlated moves of cells.
      Its graph structure is constructed by the minimum spanning tree that spans all the cell
      centroids in the initial frame.
      (b3) Determination of the cell corresponding to the root node of the tree. 
    (c) Tracking scheme at time $t$: 
      (c1) Tracking the cell corresponding to the root node by the standard particle filter.
      (c2-c4) Sequential tracking of remaining cells.
  }
  \label{fig:spf-outline}
\end{figure*}
We propose a novel tracking algorithm which we call a spatial particle filter (SPF).
Our idea to address the sampling issue of MRF-PF is to design
a better proposal distribution that shares information of cells' dynamics and
dependent moves.
{
  How do we construct such a better proposal distribution? The sampling inefficiency
  in the standard particle filter derives from the difficulty in predicting a cell 
  position based only on temporal information since moves of a nematode itself
  is sometimes sudden and rapid. The key idea to overcome the sampling inefficiency
  is to incorporate spatial information such as correlated moves among cells into
  a proposal distribution. If we already know the position of a cell near the target
  cell in the current and previous frames and the cells' moves are strongly
  correlated, the position of the target cell can be efficiently estimated from such
  correlated moves. We therefore construct a proposal distribution that predicts
  the target cell position from \textit{spatial} information, unlikely to motion
  models in the standard particle filter. 
  The remaining issue is how we determine positions of
  \textit{all} cells since all the cell positions are unknown before tracking. 
  A solution is to determine the position of a cell from a nealy-located cell, sequentially. 
}
To this end, we transform the joint distribution of
state variables for multiple targets at time $t$ into a recursive formula in 
the spaitial domain, as is similarly done in the temporal domain for the standard PF. 
Suppose the graph structure of a MRF is restricted to a \textit{tree}. 
Let $y_{t,k}$ be the region of the $k$th target image at time $t$,
$u(k)$ the parent node of node $k$, $1 \rightarrow k$ 
the path from the root node to node $k$, and $y_{1\rightarrow k}$
the set of regions of target images corresponding to nodes on the path $1 \rightarrow k$. 
Then, the conditional distribution $p(x_{t,k}|y_{1\rightarrow k})$
is represented as the following recursive formula:
\begin{align}
  &p( x_{t,k}|  y_{t,1\rightarrow k}) \nonumber\\
     &\propto p(y_{t,k}|x_{t,k}) 
          \int p (x_{t,k}|x_{t,u(k)}) p(x_{t,u(k)}|y_{t,1\rightarrow u(k)}) dx_{t,u(k)}. \nonumber
\end  {align} 
%
That is, a target can be tracked from the location of
a corresponding parent node by the standard PF-like computation for each path for each frame,
meaning all targets except for the root node can be tracked by a sweep of the MRF tree
as shown in Figure \ref{fig:spf-outline}(c).
In this formula, we design $p(x_{t,k}|x_{t,u(k)})$ as a proposal distribution
that shares information of the dynamics of the $k$th target and the energy function
between $k$ and $u(k)$.
To deal with nonlinearity and non-normality, we approximate the conditional distribution 
by a set of particles $\{x_{t,k}^{(n)}\}_{n=1}^{N}$ and 
particle weights $\{w_{t,k}^{(n)}\}_{n=1}^{N}$ as follows:
\begin{align}
  p(x_{t,k}|y_{t, 1\rightarrow k}) 
   \approx \sum_{n=1}^{N} w_{t,k}^{(n)} \delta(x_{t,k}-x_{t,k}^{(n)}).\nonumber
\end  {align} 
We note that the recursive formula is valid only when the MRF is restricted to a tree.
We therefore automatically construct an MRF tree from the initial locations of the targets,
as described later. We also note that the target corresponding to the root node in
the MRF tree cannot be tracked by this recursive formula, and so we
track the root target by the standard PF.

\subsubsection*{Detection of initial locations and construction of an MRF tree}
We detect initial cell locations of by clustering voxels with local peaks and computing 
centers of the clusters. In order to reduce computational costs, we collect voxels with 
\textit{strong} fluorescent intensities. 
Here, a \textit{strong} intensity is defined as a local maximum of fluorescent intensities 
in a region narrower than the size of a cell nucleus, 
assuming that local peaks of intensities concentrate on the center of a cell nucleus.
To cluster the voxels, we use the DP-means algorithm \cite{Kulis2012}, 
an extension of the $k$-means algorithm that is not needed to specify the number of clusters $k$.
The DP-means algorithm is suitable for our data
since (1) it automatically counts the number of clusters i.e. the number of cells, 
(2) the computation is considerably fast,
and (3) the parameter $\lambda$ that controls the cluster radii is determined
by prior knowledge of the radii of a cell. 
To construct a graph structure of a MRF that models correlated moves among cells, 
we construct a minimum spanning tree that spans all the detected locations of cell nuclei 
by Kruskal's algorithm \cite{Kruskal1956}.
The outline of the spatial particle filter is summarized in Figure \ref{fig:spf-outline}.

\subsubsection*{Particle generation in the spatial domain}
We here design a proposal distribution $p(x_{t,k}|x_{t,u(k)})$ that serves as
both the dynamics of target $k$ and the energy function between target $k$ 
and target $u(k)$. 
Our key ideas for constructing the proposal distribution are
(1) simulating the covariation with neighbor cells, 
(2) maintaining the relative positions of cells in the initial frame, and
(3) avoiding collisions with neighbor trackers.
Let $\{x_{t,k|k}^{(n)}\}_{n=1}^{N}$ be the ensemble set corresponding to
the filtering distribution $p(x_{t,k}|y_{1\rightarrow k})$ and 
let $\overline{x}_{t,k}$ be the mean vector averaged over the ensemble set.
We hereafter abbreviate the parent node $u(k)$ simply to $u$ if the confusion 
does not occur. One approach to taking the covariation information into account is
to generate particle $x_{t,k}^{(n)}$ such that the following approximation holds:
\begin{align}
           {x}_{t,k}^{(n)}   - \overline{x}_{t-1,k} \approx  
           {x}_{t,u|u}^{(n)} - \overline{x}_{t-1,u},\nonumber
\end  {align} 
that is, the movement of target $k$ between two successive frames 
is roughly the same as the movement of its parent node $u$.
In fact, this is equivalent to preserving the relative position of $k$ for $u$ 
in the previous frame, since transposition of the second and third terms yields
\begin{align}
           {x}_{t,  k}^{(n)} - {x}_{t,  u|u}^{(n)} \approx  
           \overline{x}_{t-1,k} - \overline{x}_{t-1,u}.\nonumber
\end  {align} 
The covariation of cells is utilized by this approach, 
but relative positions of cells in the initial frame are not conserved, 
i.e., the distance between a pair of cells can be infinite in the long run.
We therefore consider another constraint on particle $x_{t,k}^{(n)}$, 
such that the relative positions of cells in the initial frame are roughly conserved. 
\begin{align}
           {x}_{t,k}^{(n)} - {x}_{t,u|u}^{(n)} \approx  
           \overline{x}_{1,k} - \overline{x}_{1,u}. \nonumber
\end  {align}
We note relative positions to be conserved do not have to be those in 
the initial frame and can be arbitrarily chosen from those in all frames
if the tracking is offline. Therefore, if we find the frame when 
the movement of a nematode itself is slower and more stable than that in the initial frame,
the frame is a better choice for defining the relative positions of the cells. 
Let $\eta_t^{k_1,k_2}:=\overline{x}_{t,k_1}-\overline{x}_{t,k_2} $. 
We construct a proposal distribution $p(x_{t,k}|x_{t,u(k)})$ such that the above two 
constraints simultaneously hold:
\begin{align}
    x_{t,k}^{(n)} = x_{t,u|u}^{(n)}  +   \alpha  \eta_{t-1}^{k,u}  
                          +(1-\alpha) \eta_1    ^{k,u} + v_{t,k}^{(n)}, 
    \label{eq:motion}
\end  {align}
where $v_{t,k}^{(n)}$ is the noise vector that follows the normal distribution
with mean vector $0$ and covariance matrix $\Sigma$, and $\alpha$ 
is a tuning parameter whose range is $0<\alpha<1$ and controls the importance 
between the covariation and relative positions. 
By this construction, relative positions of cells in the initial frame are 
conserved in the long run since the time series generated from 
the proposal distribution is stationary. We see this fact by taking the average 
of Equation (\ref{eq:motion}) over the index set of particles $\{n|1 \cdots, N\}$ 
derives the following autoregressive model:
\begin{align}
    (\eta_{t}^{k,u} - \eta_{1}^{k,u})
                      = \alpha  (\eta_{t-1}^{k,u}- \eta_{1}^{k,u}) + 
                        \overline{v}_{t,k}, \nonumber
\end  {align}
where $\overline{v}_{t,k}$ is the average of the noise vectors over
the index set of particles. 
Since the autoregressive coefficient $\alpha$ satisfies $0 <\alpha <1$,
$\eta_t^{k,u}$ is a stationary process with mean $\eta_1^{k,u}$,
meaning that relative positions of cells in the initial frame are conserved in the 
long run.

The remaining task to construct the procedure of generating
particles is to avoid collisions with neighbor trackers.
To this end, we combine a rejection sampling with the proposal distribution. 
After generating a particle based on the proposal distribution, we reject the 
particle if the distance between $k$ and $u$ is small compared with the radii 
of a cell nucleus. 
Formally, we reject the particle with probability proportional to 
$\exp\{-||\eta_t^{k,u}||^2/\lambda^2\}$, where $\lambda$ denotes 
the predefined radius of cell nuclei, and accept otherwise.

\subsubsection*{Calculation of particle weights}
We next describe how we evaluate the importance of a particle which tracks a cell nucleus.
The point-spread function (PSF) is typically used for evaluating the similarity
of objects with ellipsoidal shapes \cite{Meijering2012}. 
Cell nuclei in our motivated datasets are roughly ellipsoidal but often slightly deformed. 
To evaluate similarity between cell nuclei under the existence of the deformation,
we define the importance of a particle as a similarity between 3D subimages around  
the $k$th cell centroid in the initial and current frames.
Suppose $y_t^{(w)} (x)$ is an augmented vector which corresponds to a 3D subimage at time $t$ 
with center $x \in \mathbb{R}^3$ and window width parameter $w=(w_1,w_2,w_3) \in \mathbb{N}_0^3$.
Here, the window is defined as a set of voxels in the cuboid with center voxel including 
$x$ and edge lengths $2w_1+1$, $2w_2+1$, and $2w_3+1$. 
We also suppose that $\hat{x}_{1,k}$ is the initial position of the $k$th cell centroid 
detected by the DP-means algorithm. 
We then define a likelihood function of particle $x_{t,k}^{(n)}$ as follows:
\begin{align}
  p(y_{t,k}|x_{t,k}^{(n)}) \propto 
     \exp 
       \Bigg\{ 
        - \frac{||y_t^{(w)}(x_{t,k}^{(n)})- y_1^{(w)}(\hat{x}_{1,k}) ||^2}{2\sigma^2 W} 
       \Bigg\}, \nonumber
\end  {align}
where $\sigma^2$ is the parameter that controls the variance of the 3D-subimage similarity
and $W$ is the number of non-zero elements for the vector defined as the elementwise sum 
of $y_t^{(w)}(x_{t,k}^{(n)})$ and $y_1^{(w)}(x_{t,k}^{(n)})$.
We note that the difference is averaged over nonzero elements in the elementwise sum of 
the subimages in order to avoid an unintended similarity increase derived from voxels 
within the window but outside the nucleus region.  We also note that the window parameter 
$w$ should be chosen to be larger than the maximum size of cell nuclei so that
all the information of shapes and intensities are included in the window. 

\subsubsection*{Dealing with the disappearance of targets to be tracked} 
Another difficulty in tracking cells is that 
some neurons go out of the image space because of movement of the nematode.
We track cells out of the image space by the following scheme.
If a generated particle in the prediction step goes out of the image space, 
we add the particle to a member of the filter ensemble without resampling. 
The reason why we skip the filtering step for such a particle is that likelihood 
of the particle cannot be evaluated 
whereas the particle should be survived considering the possibility that the 
target truly goes out of the image space.  

\subsubsection*{Software}
We developed a software suite SPF-CellTracker. Our software is composed 
of three executables, \textit{convert}, \textit{track}, and \textit{view}.
The first software \textit{convert} converts a set of 2D images that compose 4D live-cell imaging 
data into a single file encoded as our original binary format. 
During conversion, the average subtraction and the 3D median filter can be optionally 
applied for each 3D image in order to remove background noise and salt-and-pepper noise.
The second software \textit{track} is the main software for detecting and tracking multiple cells 
based on the SPF algorithm from the converted 4D image file. 
This software was implemented purely in the C language considering the running speed,
which will be presented in the following section. The third software \textit{view} is to 
visualize the 4D image data with a tracking result. An interesting feature of this visualization
software is an ability to zoom in, zoom out, and rotate 4D image during playing 4D image data. 
All executables and source codes of \textit{convert} and \textit{track} are freely available in 
the supplementary website (S1) listed in Appendix.

\section{Experiments}
  We here present comparisons of cell-tracking and cell-detection performance using 
  synthetic data and real 4D live-cell imaging data. 
  All of supplementary videos described in this section are available in 
  the supplementary website (S2) listed in Appendix.

\subsection{Comparison of tracking performance}

\subsubsection*{Application to synthetic data}
 We begin with how we generated synthetic datasets.
 We constructed a model that simulates real 4D imaging data with following characteristics:
 (1) relative positions of cells obtained by real 4D imaging data are preserved in the long run, 
 (2) movements of nearly-located cells strongly correlate, 
 (3) cells are imaged as globular-like objects and it is difficult to discriminate cells by shape and size, 
 and (4) cells do not divide but occasionally disappear.
 We assumed the use of confocal microscopes for imaging cell nuclei. 
 Typical confocal microscopes generate 3D images with a longer step size 
 along $z$-axis than the edge length of a pixel in $xy$-plane.  
 We therefore set the step size along $z$-axis is three times larger than 
 those in $xy$-plane. To be reasonably consistent with our real 4D live-cell imaging data, 
 we set the resolution and the number of frames to $512\times 256\times 20$ and $500$, respectively. 
 We used the results of cell detection by the DP-means algorithm as initial positions of cells 
 for each of three real 4D imaging data. For the three datasets, $114$, $120$ and $115$
 cells were detected. We used the prediction model in SPF as a simulation model of cells'
 correlated moves. Locations of the root nucleus after the second frame were not changed 
 in order to stabilize the position of a simulated nematode itself.
 Positions of remaining non-root nuclei after the second frame were generated by Equation 
 (\ref{eq:motion}) with $\alpha=0.6$ and $\Sigma^{-1/2}=\textrm{diag}(0.6,0.6,0.03)$ 
 to simulate correlated moves of nearly-located cells and preservation of relative positions among cells. 
 These parameters were determined based on the visual similarity between generated data and real imaging data 
 through trial and error. Especially, we set the variance of noise along $z$-axis to nearly zero since 
 a move of a nematode in the dorsal-ventral axis is considerably small even if a move in the anteroposterior 
 axis is relatively large.  
 To generate globular-like objects as images of cells, we used an anisotropic point-spread function (PSF).
 Suppose $d_{\Lambda}(x,\mu)^2=(x-\mu)^T \Lambda^{-1} (x-\mu)$ is the square of the Mahalanobis 
 distance between $x\in \mathbb{R}^3$ and $\mu\in\mathbb{R}^3$ with covariance matrix 
 $\Lambda\in\mathbb{R}^{3\times3}$. 
 We define a PSF with nucleus center $\mu$ and shape parameter $\Lambda$ as follows:
  \begin{align}
  f(x;\mu,\Lambda) \nonumber 
  =\begin{cases}
  c \exp \big\{-\frac{1}{2} d_{\Lambda}(x,\mu)^2 \big\} & \text{if }  d_{\Lambda}(x,\mu) <1 \\ 
                          0                       & \text{otherwise},
  \end  {cases} \nonumber
  \end  {align} %
  where $c$ denotes intensity at the nucleus center $\mu$.
  For all nuclei, we used the same intensity parameter $c$ and shape parameter $\Lambda$, 
  leading to the most severe condition to track since cell nuclei are not distinguishable from
  the intensity and shape information. We set the shape parameter $\Lambda=\text{diag}(9,6,3)$, 
  which corresponds to an ellipsoid where its principal axes are parallel to the $x$, $y$, and $z$ axes 
  and those lengths are $9, 6$, and $3$ voxels. A nucleus image was deleted with probability $0.03$ 
  for each nucleus for each frame in order to simulate its disappearance.
  Supplementary video 1, 2 and 3 show resulting synthetic datasets. The videos show similar motions
  of cell nuclei in real datasets although the large deformation of a nematode body such as S-curve 
  is not reproduced. 

\begin{figure}[]
   \centerline{ 
   \includegraphics[scale=0.7]{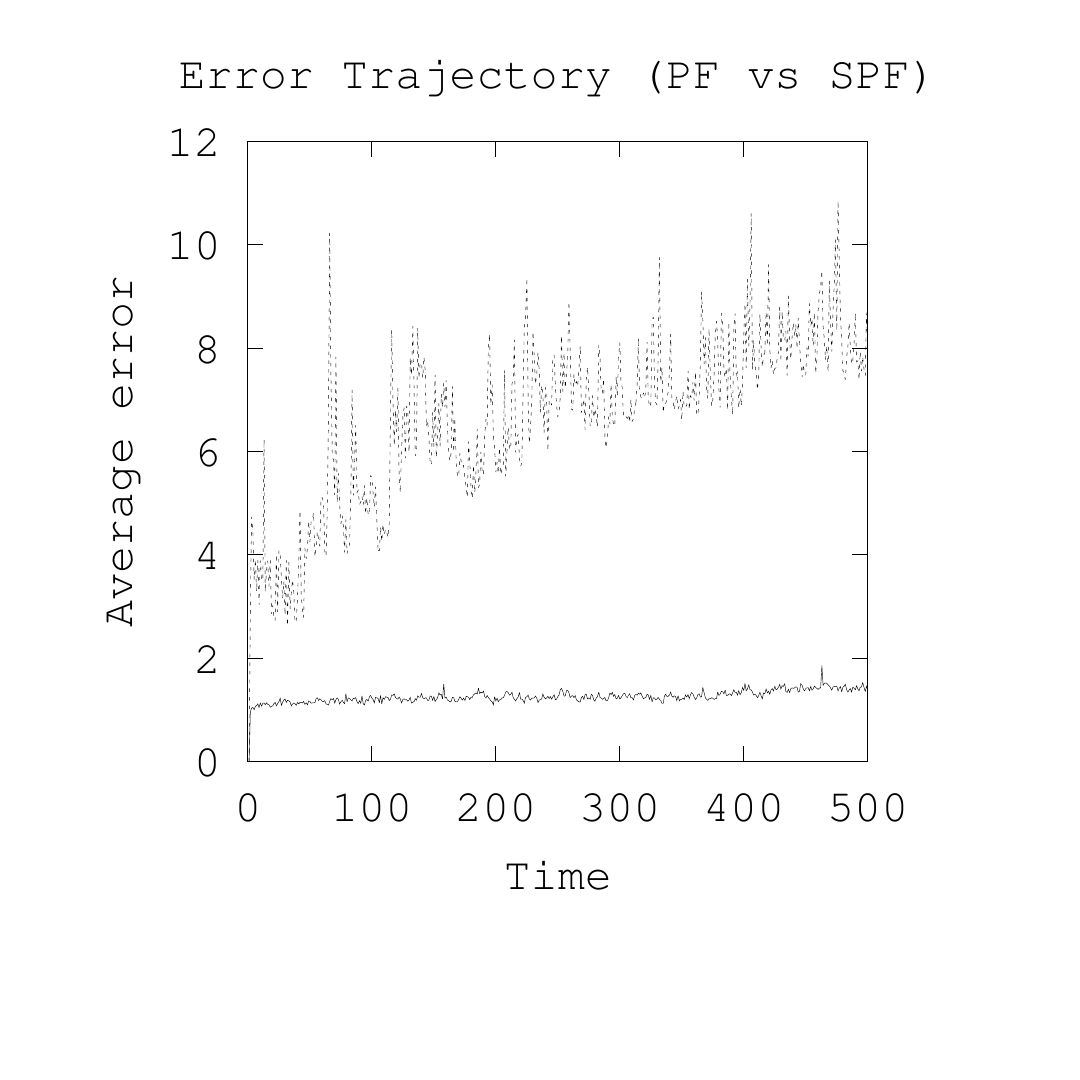} 
   }\vspace{-15mm}
   \caption{
            Time series plots of average tracking errors per cell nucleus per trial of 
            the standard PF (dotted line) and SPF (solid line) for the synthetic dataset 1. 
            Errors were calculated as the 
            root mean square error (RMSE) from true positions of synthetic nuclei.
            The unit distance was defined as the edge length of a voxel in $xy$-plane.
            The error along $z$-axis was expanded to three times of those in $xy$-plane 
            considering the difference in physical length between $xy$-plane and $z$-axis.
           }
   \label{fig:traj}
 \end{figure}

 Next, we proceed to the evaluation of tracking performance for the synthetic datasets described above.
 For the three datasets, we conducted 20 trials for each PF and SPF.
 We set the same parameters to those used for generating the synthetic image data.
 We then computed root-mean-square errors (RMSEs) and counted the number of tracking failures. 
 In computing RMSEs, distance along $z$-axis was expanded to three times its initial
 length since the physical length of a voxel along $z$-axis is assumed to be three times larger 
 than those in $xy$-plane in order to simulate anisotropy of the 3D imaging.
 Figure \ref{fig:traj} shows time series plots of the average RMSE of PF and SPF
 per nucleus per trial for the synthetic dataset 1. The error of PF gradually increases as 
 time $t$ increases, while the error of SPF is considerably small and stable. 
 Table \ref{tab:sim} summarizes the failure counts of whole applications to the three datasets. 
 The tracking failure was defined as the case that the distance between the true position and its 
 estimate is greater than the radius of the synthetic nucleus, \textit{i.e.},
 $4.5$ voxels. The failure count was \textit{not} incremented if a tracking failure 
 was inherited by the next frame, but it \textit{was} incremented if the tracking failure 
 occurred again after the recovery from the tracking failure.  
 As shown, SPF drastically reduced the number of tracking failures in comparison with 
 those of PF. 
 \begin{table}[]
 \caption{The average number of tracking failures per nucleus per trial in 500 frames. 
          PF and SPF were applied to three synthetic datasets.
          The numbers of tracking failures were averaged over the number of nuclei and 
          the number of trials. Estimated standard errors are shown in parentheses. 
          }
   \begin{center}
     \begin{tabular}{ccc} \hline 
        Data    & PF              & SPF           \\ \hline
        1       & 37.36 (0.234)   & 1.43 (0.038)  \\
        2       & 41.60 (0.179)   & 0.78 (0.016)  \\
        3       & 37.22 (0.218)   & 1.14 (0.013)  \\ \hline
     \end{tabular}
   \end{center}
   \label{tab:sim}
 \end{table}

\subsubsection*{Application to real 4D live-cell imaging data}
  We evaluate tracking performance of SPF and the state-of-the-art method reported by 
  Tokunaga {\em et al.}\cite{Tokunaga2014}. The data we used are D1, D2 and D3 in Data II reported by them,  
  that is, 4D live-cell imaging data obtained by imaging nuclei of \textit{C. elegans} neurons
  (Supplementary video 4-6).
  Each of the datasets is composed of 500 frames of 3D images with resolution $512\times 256\times 20$, 
  \textit{i.e.,} 20 $z$-slices of 2D images with resolution $512\times 256$. 
  Since the background noise level of the datasets changes according to the depth of the 3D image, 
  we approximately estimated the background noise level as the average over fluorescent intensities 
  of a 2D image and subtracted it from the 2D image. We then applied the median filter
  with window size $3\times 3\times 1$ to remove the salt-and-pepper noise.
  To obtain tracking results independent from the quality of cell detection, we used the
  same starting positions for the both methods. The starting positions were obtained
  by using the repulsive hill-climbing (RPHC) algorithm \cite{Tokunaga2014}.
  All the parameters required for SPF were manually-tuned and the same parameters were
  used for all of the three imaging data. The parameter set used for SPF and
  the resulting tracking animations of SPF (Supplementary video 7-9) and \cite{Tokunaga2014} 
  (Supplementary video 10-12) are available in the supplementary website.
  Figure \ref{fig:compare-f} shows tracking results of the both methods in the final frame of D3 in Data II. 
  The top panel of the figure shows the starting positions of trackers.
  Five trackers that do not track cell centroids in the bottom of the image space are 
  derived from false positives of cell-detection in the initial frame. 
  The middle and bottom panels in Figure \ref{fig:compare-f} show positions of trackers
  in the final frame for \cite{Tokunaga2014} and SPF, respectively. 
  The figure suggests that at least large inconsistencies between cell centroids and 
  positions of trackers for the both methods do not exist.

  We evaluated tracking performance using the final frame of original datasets
  by manual verification. To reduce errors of manual verification as much as possible, 
  we implemented following features for our 4D image viewer: 
  (1) focusing only one cell and the corresponding tracker, 
  (2) starting and stopping at an arbitrary frame,
  (3) rotation of the 4D image, and
  (4) adjustment of the intensity threshold that determines whether or not a voxel is displayed.
  These features of the 4D image viewer are introduced in Supplementary video 13.
  By using the features of our 4D image viewer, we calculated success rates of the both methods
  based on the following principles:
  \begin{itemize}
  \item { A tracker was counted as \textit{success} if the distance from the corresponding cell centroid
    was within $5$ voxels.  }
  \item { A tracker was excluded from the calculation if it was an incorrectly detected cell centroid
    in the initial frame.}
  \item { A tracker was also excluded from the calculation if the corresponding cell centroid was
    out of view in the final frame.  }
  \end  {itemize}
  Table \ref{tab:real} shows the success rates of the both methods based on the manual verification. 
  SPF considerably improved tracking performance comparing with \cite{Tokunaga2014} for all
  the three datasets D1-D3 in Data II. Among the three datasets, tracking performance in D3 was noticeably
  lower than the other datasets for the both methods. This derives from the differences of nematodes'
  body postures and movements in the final frame. The body postures in the final frame of D1 and D2 are
  roughly straight and movements of the nematode are moderate (Supplementary video 4 and 5).
  On the contrary, the nematode in the final frame of D3 is largely shrunk along the
  anterior-posterior axis and the nematode's movement is rapid (Supplementary video 6).
\begin{figure}[]
  \centerline{ \includegraphics[scale=0.35]{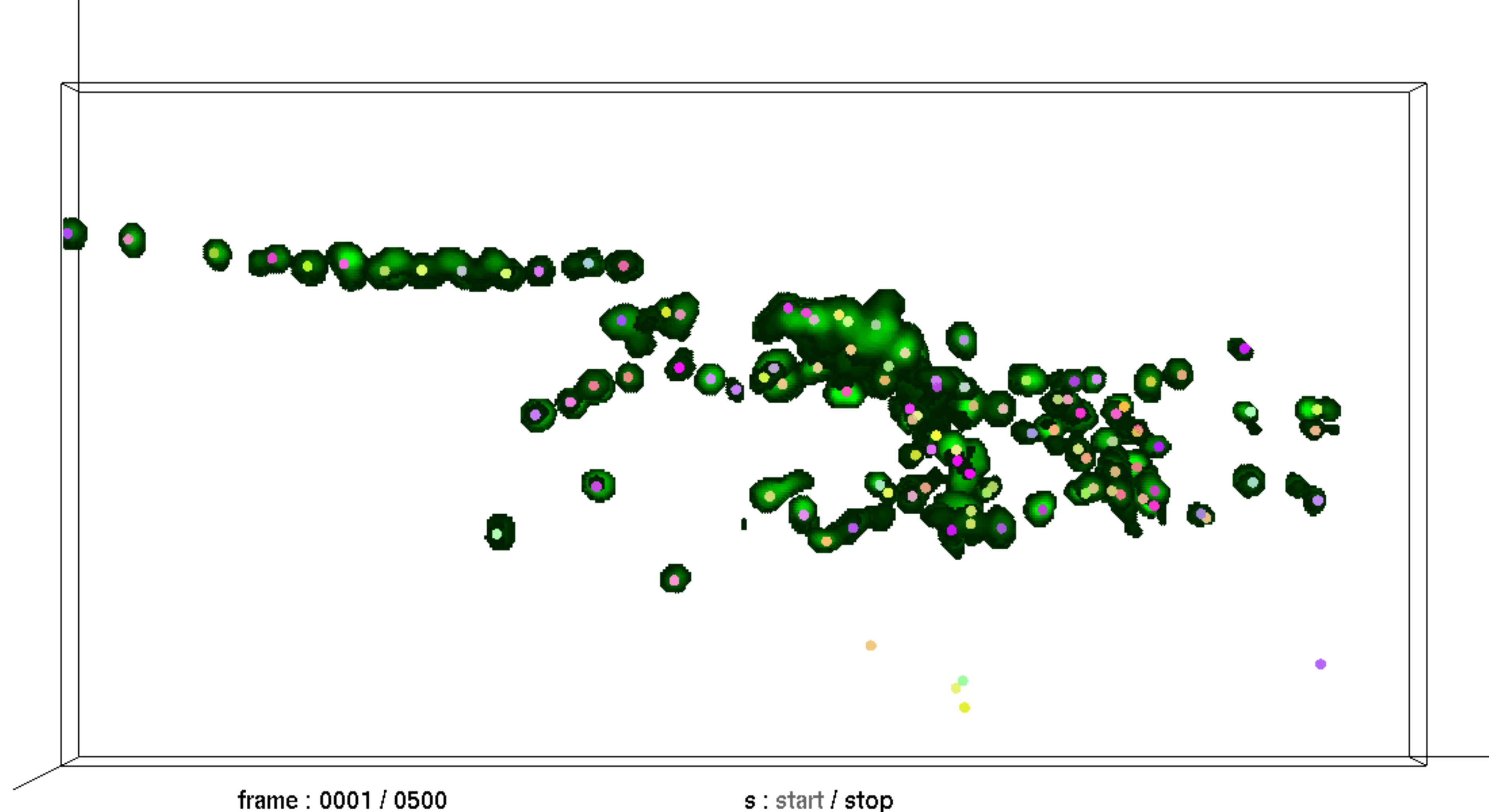} }
  \centerline{ \includegraphics[scale=0.35]{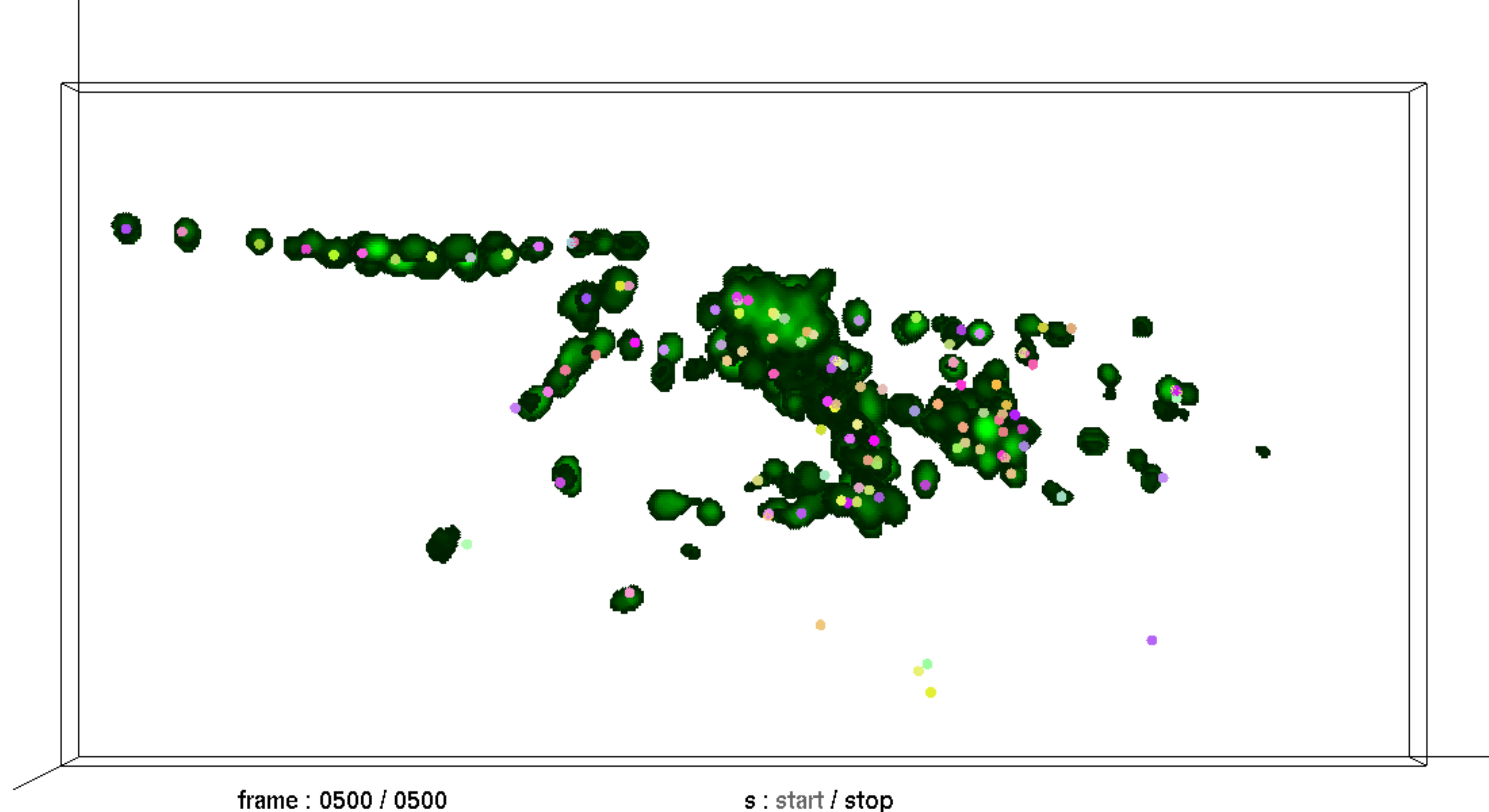} }
  \centerline{ \includegraphics[scale=0.35]{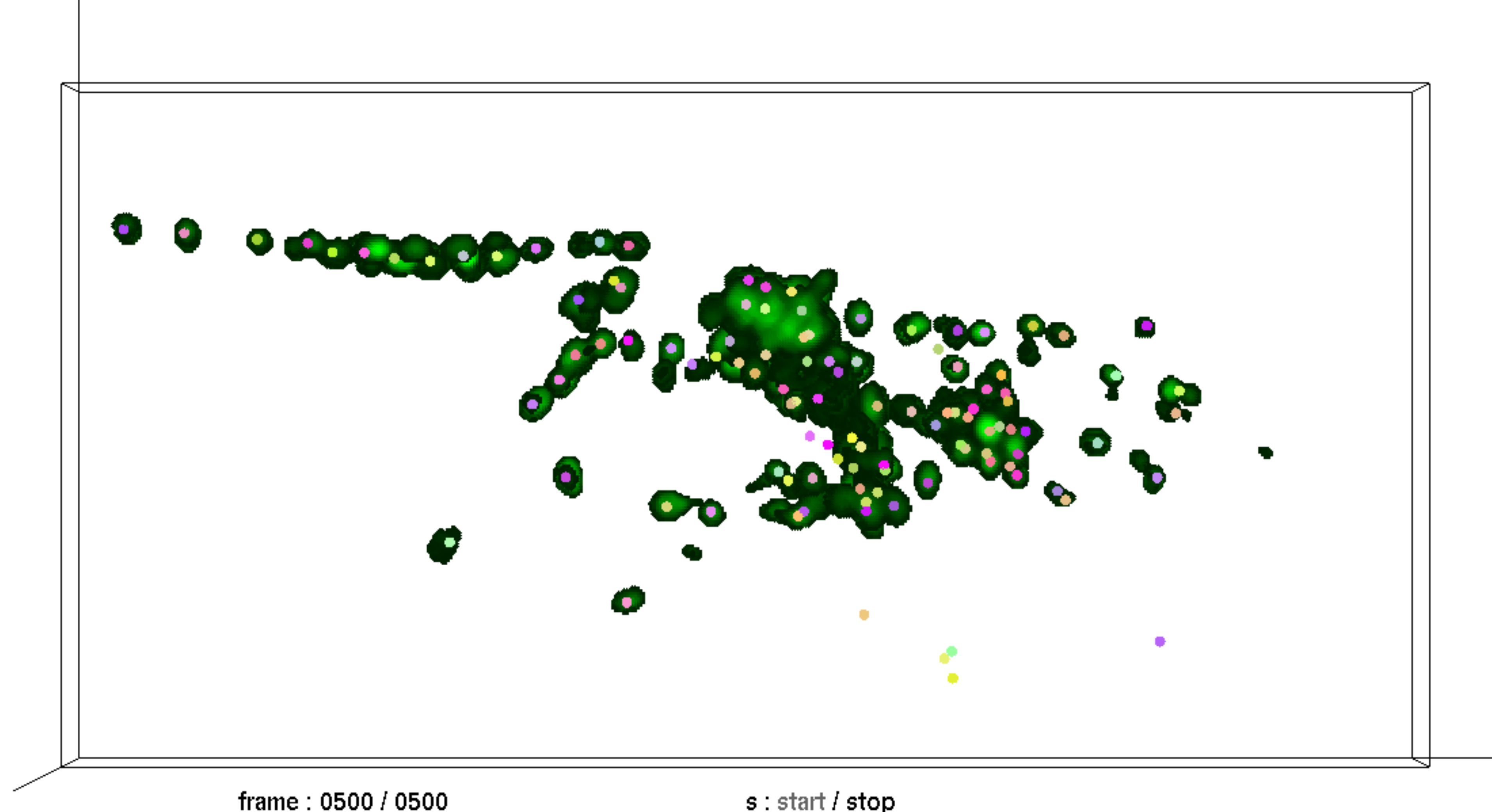} }
  \caption{Comparison of tracking performance for D3 in DATA II. 
    (Top) Starting positions of trackers for the both methods. Trackers are depicted by colored circles.
    (Middle) Tracking result of \cite{Tokunaga2014} in the final frame.
    (Bottom) Tracking result of SPF in the final frame.
  }
  \label{fig:compare-f}
\end{figure}
\begin{table}[]
\caption{
    The rate of trackers which correctly tracked corresponding cells in the final frame $t=500$ 
    for D1-D3 in Data II. The number of cells in the initial frame detected by the RPHC algorithm is
    shown in parentheses of the first column. Consistency of tracking between the initial 
    and final frames was manually checked for each cell using our 4D image viewer.
  }
  \begin{center}
    \begin{tabular}{cccc}\hline
       Data      &  Tokunaga \textit{et al.} (2014)  & SPF       \\ \hline
       D1 (120)  &  0.8000                           & 0.9524    \\
       D2 (124)  &  0.6860                           & 0.9669    \\
       D3 (111)  &  0.4467                           & 0.8679    \\ \hline
    \end{tabular}
  \end{center}
  \label{tab:real}
\end{table}

\subsection{Comparison of cell-detection performance} 

\begin{figure}[] 
  \centerline{ 
    \includegraphics[scale=0.38]{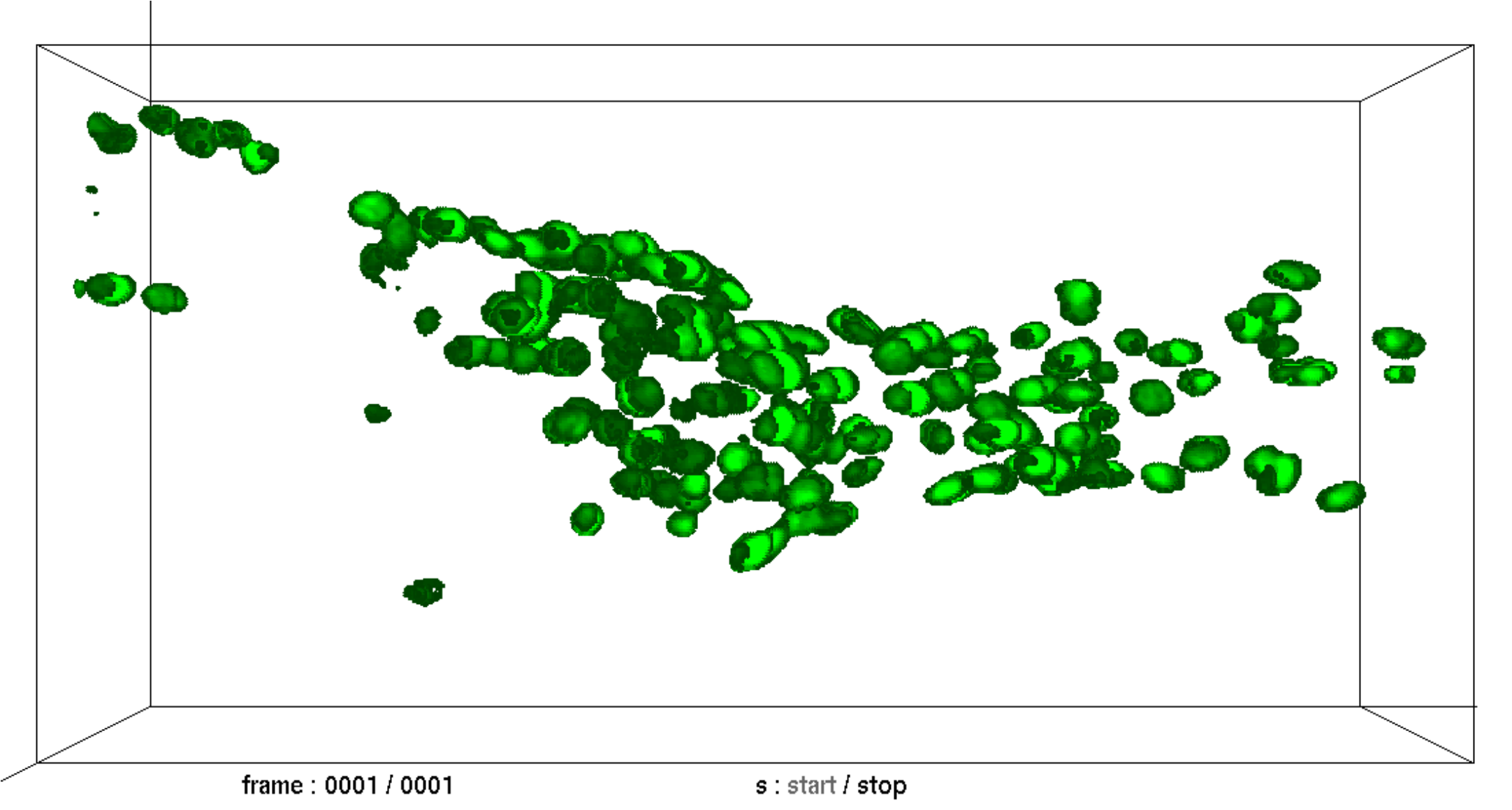} 
  }\vspace{3mm}
  \caption{
    A 3D still image in DATA I that captured nuclei of \textit{C. elegans} 
    neurons after noise removal.
  }
  \label{fig:data1}
\end{figure}

\begin{figure}[] 
  \centerline{ 
    \includegraphics[scale=0.28]{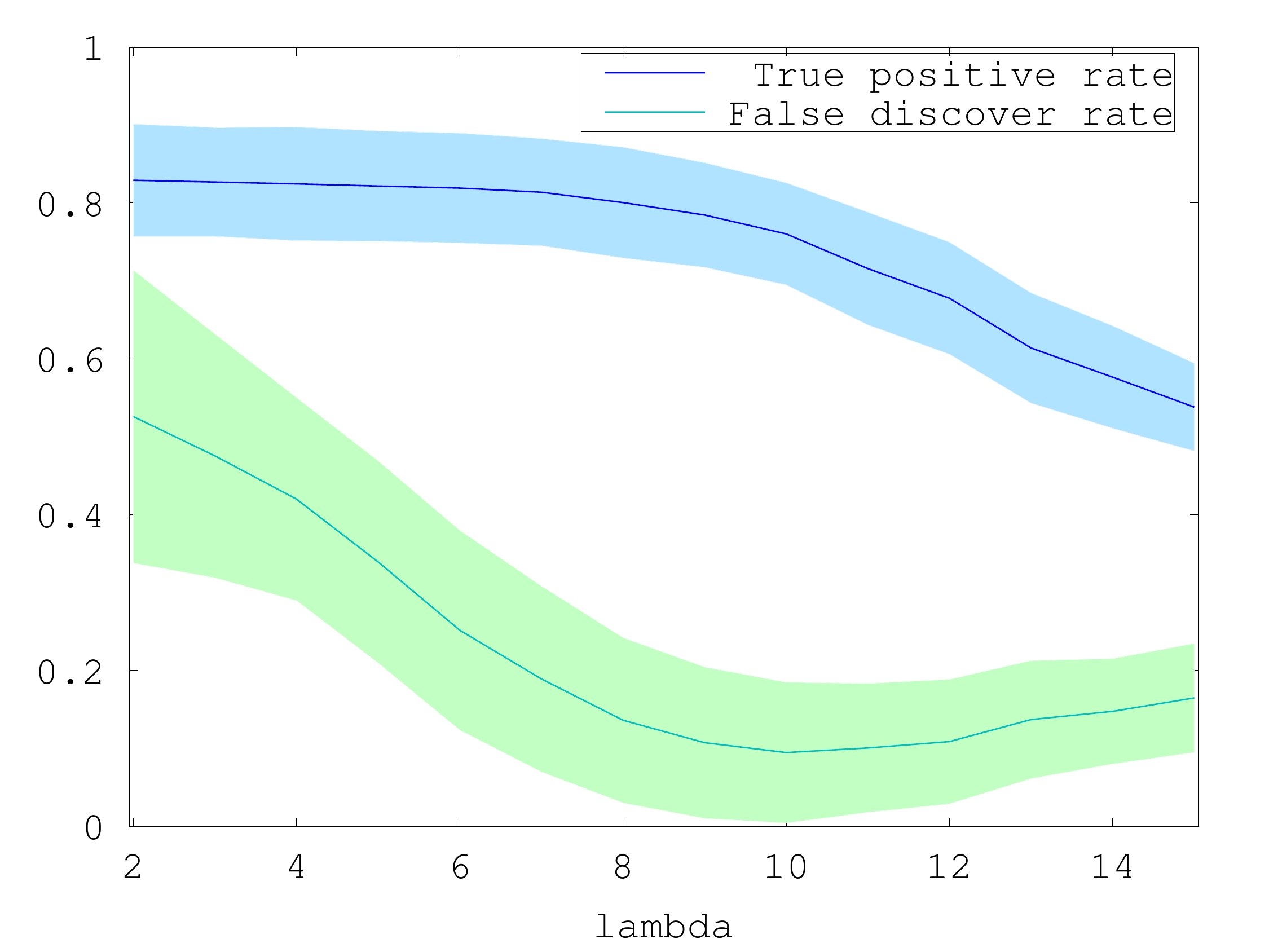} 
  }\vspace{-3mm}
  \caption{
    The radius parameter $\lambda$ and cell-detection performance of the DP-means algorithm. 
    The accuracy was averaged over results of ten 3D still images in DATA I. 
    Ranges within one standard deviation are indicated by shaded regions.
  }
  \label{fig:lambda}
\end{figure}

  Next, we evaluate detection performance of the DP-means algorithm by comparing that of 
  RPHC algorithm \cite{Tokunaga2014}. The datasets we use are DATA I reported by them.
  DATA I was obtained by imaging nuclei of nematode's neurons likely to DATA II
  but is different in that 
  (1) it is composed of distinct 3D still images instead of 4D images and
  (2) cell centroids were manually annotated in order to obtain the ground truth of 
  the cell-detection problem.
  The resolution of $xy$-plane of all the ten datasets is $512\times 256$ and
  the numbers of $z$-slices for the datasets range from $119$ to $203$.
  A 3D image in Data I after the noise removal is shown in Figure \ref{fig:data1}.
  To evaluate the performance of the both algorithm,
  we counted the number of true positives (TP), the number of false positives (FP), 
  and the number of false negatives (FN).
  A position detected by an algorithm is called a true positive if the position 
  is within 5 voxels from a manually identified position, and is called a false positive 
  otherwise. An annotated position that is not detected by an algorithm is
  called a false negative. We then computed true positive rates and false discovery rates 
  defined as TP/(TP+FN) and FP/(FP+TP), respectively.
  Table \ref{tab:detect} shows false discovery rates and true positive rates of the 
  both algorithms. The radius parameter $\lambda$ of the DP-means algorithm was set to $8$ 
  voxels close to the maximum radius of cells. 
  The true positive rate of the DP-means algorithm is roughly the same as RPHC 
  while its false discovery rate were higher than RPHC. 
  Figure \ref{fig:lambda} shows an effect of the radius parameter $\lambda$ on cell-detection
  performance of the DP-means algorithm, suggesting that the DP-means algorithm achieved
  the best for $\lambda=8$ and $\lambda=9$ that are close to the maximum size of cells.

\begin{table}[]
\caption{
    False discovery rate and true positive rate of cell detection.
  }
  \begin{center}
    \begin{tabular}{lccc}\hline
                                  &  RPHC            & DP-means ($\lambda=8$) \\ \hline
       True positive rate         &  0.8041 (0.0362) & 0.8004 (0.0710)        \\ %
       False discovery rate       &  0.0301 (0.0305) & 0.1362 (0.1060)        \\ \hline
    \end{tabular}
  \end{center}
  \label{tab:detect}
\end{table}

\subsection{Computational Time}
\begin{figure}[t]\vspace{-1cm} 
  \centerline{ 
  \includegraphics[scale=0.35]{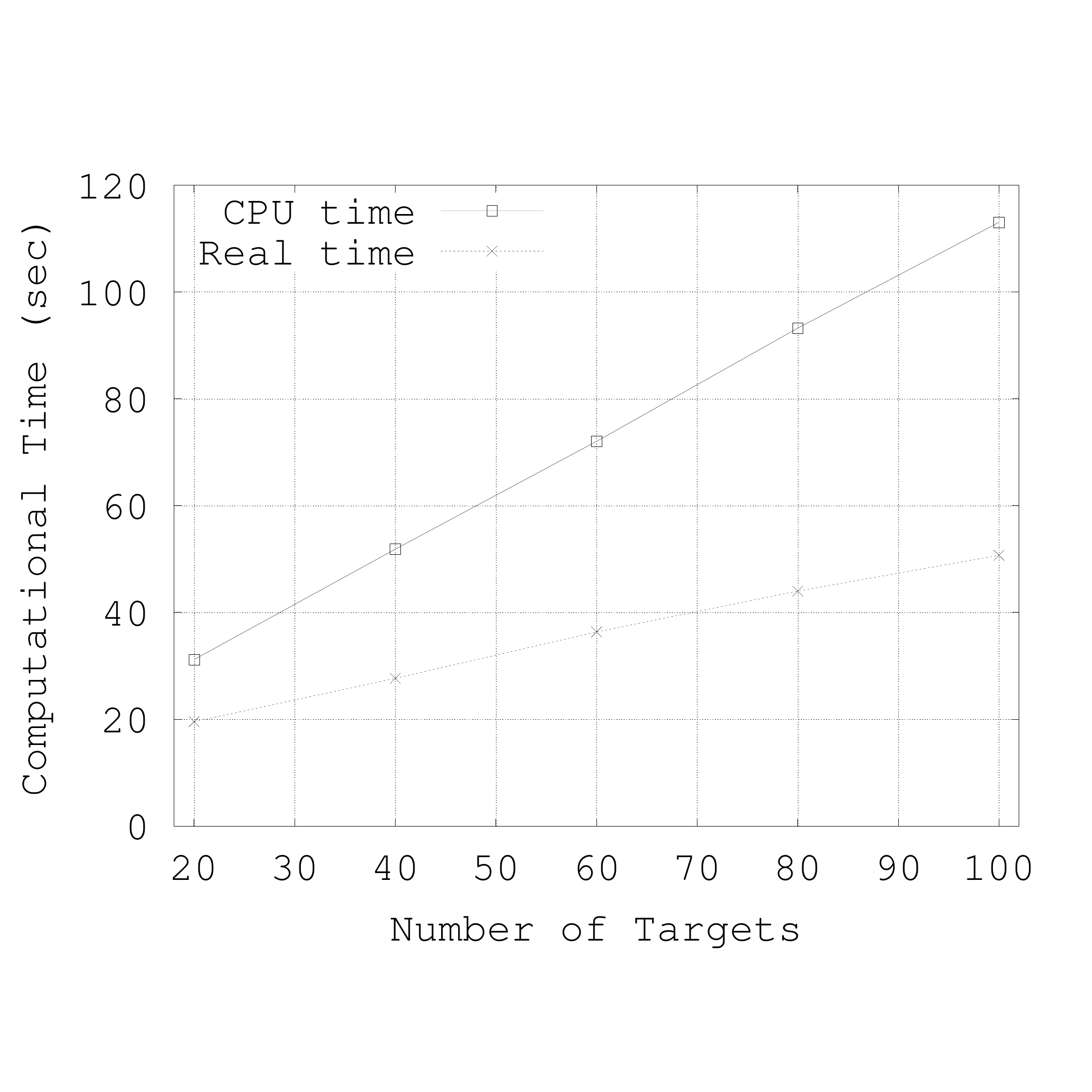}
  }\vspace{-1cm} 
  \caption{
   The average time consumption of SPF and its scalability to the number of targets. CPU time and real 
   time were evaluated for completing tracking of all $500$ frames of the dataset D1 in DATA II. 
  }
  \label{fig:time}
\end{figure}

  We computed CPU time and real time for D1 in DATA II that is composed of $500$ frames of 
  3D images with resolution $512 \times 256\times 20$. We used iMac ($27$-inch, Late $2013$,
  OS X $10.9.5$) with $3.2$GHz Intel Core i$5$ and $8$GB RAM as our computational environment. 
  OpenMP implemented in GCC 4.8 was used as a tool of parallelization. 
  We used the same parameter set as that used in the evaluation of tracking performance for 
  real 4D live-cell imaging data. Especially, we set the number of particles for tracking a 
  cell to $1000$, i.e. the total number of particles for tracking $100$ cells is $10^5$. 
  To measure the scalability of our method, 
  we randomly selected $20,40,60,80$, and $100$ cells without replacement among $114$ cells. 
  Figure \ref{fig:time} shows CPU time and real time consumed for tracking 
  $500$ frames in the dataset D1 in DATA II. The CPU time and real time was averaged over $20$ 
  trials.
  The CPU time in the most severe condition i.e. $K=100$ was less than 120 seconds, 
  suggesting that our software is sufficiently of practical use.

\subsubsection*{Discussion}
We here summarize the results suggested by the numerical studies.
\begin{itemize}
\item The tracking performance of the standard PF was substantially improved by the proposed method 
      for synthetic datasets (Table \ref{tab:sim}).
\item The tracking errors of PF were sequentially inherited by the next frame, while that of SPF 
      was not severely inherited. This result suggests a recovery from tracking failures 
      (Figure \ref{fig:traj}).
  \item Tracking performance can be evaluated by using the features implemented in our 4D image viewer
        (Supplementary video 13). 
  \item SPF considerably outperformed Tokunaga \textit{et al.} in terms of success rates of tracking 
        in the final frame for all the three datasets D1-D3 in Data II (Table \ref{tab:real}).
  \item Parameters of SPF were set to the same ones for all the three datasets D1-D3 in DATA II,
        suggesting a robustness of SPF (Section 3.1).
  \item For the method of Tokunaga \textit{et al.}, some trackers were merged with other trackers
        and did not return to cells to be tracked again (Supplementary video 10-12).
  \item For SPF, some trackers often recovered from tracking failures and merged trackers were rarely observed, 
        probably due to the integration of spatial information such as covariation, relative positions, 
        and collision avoidance among cells (Supplementary video 7-9).
  \item True positive rates of the DP-means algorithm and the RPHC algorithm for detecting cell-centroids
        were roughly the same while the false positive rate of the DP-means algorithm was higher than that 
        of the RPHC algorithm.
        This suggests an advantage of the RPHC algorithm over the DP-means algorithm for the detection of
        cell centroids (Table \ref{tab:detect}).
  \item The CPU times of SPF was proportional to the number of targets,
        suggesting a favorable property in terms of the scalability to the data size.
        (Figure \ref{fig:time}).
\end{itemize}

\section{Conclusion}
In this study, we aimed at tracking of more than a hundred of cells in 4D live-cell imaging data. 
One important characteristic of live-cell imaging data is that cells to be tracked are densely 
scattered and visually similar.
For these imaging data, the particle filter often mistakes a cell of interest for 
the other cells, since visual similarity among cell nuclei makes their discrimination difficult.
Fortunately, our 4D live-cell imaging data shares a characteristic that is useful 
for accurate tracking: 
cells' moves in the 4D live-cell imaging data are strongly correlated
and the relative positions among target cells are roughly conserved.

To address the tracking issue, we designed an MRF
that models the covariation and preservation of relative positions among 
cells. 
To avoid the inefficiency of JPF and MCMC sampling, we also proposed a novel 
sampling algorithm which we call spatial particle filter. The proposal distribution 
in the prediction step draws more accurate particles than those generated
by dynamics since the proposal distribution shares both temporal and spatial 
information of a cell's moves. 
SPF tracks cells by a sweep of an MRF tree in the spatial order for each frame; 
this allows effective simultaneous tracking. The MRF tree is automatically constructed
by computing an MST among the initial locations of all targets. 
We applied the proposed method to synthetic data and our 4D live-cell imaging data
of \textit{C. elegans}. The results showed that our advantage over our previous
algorithm. 

Future work includes performance comparisons with tracking methods
such as detection-and-linking and contour-evolution methods.
Another direction is to expand the applicability of SPF to a wider class of 4D 
imaging data,
for example, 4D images of chromosome arrangements during cell division. Tracking 
chromosomes is a challenging problem since they duplicate, split, and change
their shapes according to the state of cell division. 
We expect the proposed method to be a reasonable candidate for tracking multiple 
targets in a wider class of 4D live-cell images.

\section*{Appendix}
The source codes of SPF-CellTracker are available at the 
following github repository (S1).  All the supplementary videos 1-13
described in Expeiriments section are available at the following 
supplementary website (S2).  
\begin{tabular}{ll} 
 \hspace{-1mm}(S1) & \hspace{-4mm} https://github.com/ohirose/spf \\
 \hspace{-1mm}(S2) & \hspace{-4mm} https://sites.google.com/site/webosamuhirose/spf 
\end{tabular}

\section*{Conflict of interest}
None declared.

\section*{Acknowledgment}
This study is supported in part by the CREST program ``Creation of
Fundamental Technologies for Understanding and Control of Biosystem
Dynamics'' of Japan Science and Technology Agency (JST).

\end  {document}